\documentclass[11pt,twocolumn,twoside]{article}
\usepackage{fully3d}
\usepackage{graphicx}
\usepackage{booktabs}
\usepackage{tabularx}

\begin{document}

\title{GB-SVFBP: Gaussian-Based Shift-Variant FBP neural network} 


\author{Chengze~Ye}
\author{Linda-Sophie~Schneider}
\author{Yipeng~Sun}
\author{Andreas~Maier}

\affil{Pattern Recognition Lab, Friedrich-Alexander University Erlangen-Nuremberg, Erlangen, Germany}


\maketitle
\thispagestyle{fancy}


\begin{customabstract}

This paper proposes a Gaussian-Based Shift-Variant filtered backprojection (FBP) neural network, which is designed for the efficient reconstruction of non-circular trajectory cone beam computed tomography. The traditional differentiable shift-variant FBP model consists of a filtering component and a backprojection process. The filtering component includes operations such as weightings, differentiations, a 2D Radon transform, and a 2D backprojection. The proposed methods build on this framework by introducing a trainable 2D Gaussian model to represent the trajectory-related part in the filtering process, achieving a substantial reduction in the number of trainable parameters. Experimental results demonstrate that the proposed model reduces the parameter count by 99\%, while only sacrificing a slight amount of reconstruction quality. Furthermore, the training time for each trajectory is reduced to one-fourth of the original, significantly accelerating convergence. These enhancements demonstrate a considerable augmentation in the model's practicality and effectiveness, making it a valuable asset for real-world applications.
\bigskip


\end{customabstract}


\section{Introduction}

Cone beam computed tomography (CBCT) has become an essential imaging modality in fields such as medical diagnostics, industrial inspection, and security screening. Its ability to provide three-dimensional (3D) imaging with high spatial resolution makes it an attractive choice for a wide range of applications. Traditionally, CBCT systems have relied on circular scanning trajectories due to their simplicity. However, there has been a recent surge of interest in exploring non-circular trajectories, such as sinusoidal trajectories, due to their unique advantages. These trajectories can provide enhanced coverage of the region of interest, minimize artifacts caused by metal objects~\cite{herl2020scanning}, and facilitate more adaptable system designs suitable for constrained environments or specialized imaging requirements.

Despite these benefits, the adoption of non-circular trajectories introduces new challenges for reconstruction algorithms. The conventional filtered back-projection (FBP) method, which is widely valued for its computational efficiency, assumes a circular trajectory and is therefore fundamentally unsuitable for non-circular scanning paths. In order to address this limitation, the utilization of iterative reconstruction methods has been employed. While these methods are effective in handling non-circular trajectories, they are characterized by high computational complexity, which hinders their practicality in time-sensitive or resource-constrained applications.

In order to address this limitation, Defrise and Clack~\cite{defrise1994cone} proposed a shift-variant FBP algorithm. This approach, which integrates a filtering part with a backprojection process, enables simultaneous data acquisition and reconstruction, thereby markedly enhancing reconstruction efficiency. However, it should be noted that the shift-variant FBP algorithm has its own limitations. Specifically, for each unique trajectory, a dedicated filtering component must be carefully designed, which becomes increasingly complex and challenging as the scanning paths deviate further from simple geometries.

In order to address this issue, Ye~\textit{et al.}~\cite{ye2024deep,ye2024draco} proposed a differentiable shift-variant FBP model based on known operator learning~\cite{maier2019learning}, which integrates data-driven approaches with the conventional physics-based reconstruction process. This model maps the shift-variant FBP algorithm into a neural network framework, thereby enabling data-driven estimation of the trajectory-related components in the filtering process. This approach eliminates the necessity for manually designing filters for each trajectory, thereby significantly enhancing the flexibility and adaptability of the reconstruction algorithm. However, the model's parameter count becomes exceedingly large due to the necessity of estimating a unique weight for each projection during the filtering process.

Ye~\textit{et al.}~\cite{ye2025compressibilityanalysisdifferentiableshiftvariant} demonstrates that the trained parameters in the neural network exhibit significant redundancy and compressibility when subjected to principal component analysis (PCA). By decomposing these weights into a trainable eigenvector matrix, compressed weights, and a mean vector, the method effectively compresses the network parameters, achieving a reduction of nearly 80\%. However, the linear transformation used in this approach imposes inherent constraints on the compression capability.

In order to address these constraints, a Gaussian-Based Shift-Variant FBP (GB-SVFBP) model is proposed. This network extends the differentiable FBP framework by introducing a trainable two-dimensional (2D) Gaussian Mixture Model (GMM) to represent trajectory-related components in the filtering process. This design significantly reduces the number of trainable parameters while preserving high reconstruction quality. Specifically, our approach achieves a substantial 99\% reduction in the number of trainable parameters compared to traditional differentiable shift-variant FBP models, accompanied by a minimal accuracy trade-off.

\section{Materials and Methods}

\subsection{Differentiable Shift-Variant FBP Model}
\label{sec:2.2.1}

The neural network adaptation of the shift-variant filtered backprojection (FBP) method~\cite{defrise1994cone} introduces a data-driven framework for reconstruction that extends the traditional algorithm with modern machine learning techniques.

This paradigm represents a significant development in the field by reformulating the conventional shift-variant FBP algorithm into a fully differentiable framework. This transformation is achieved by leveraging principles of known operator learning. A key advantage of this approach is its capacity to optimize the trajectory-related part during filtering through backpropagation, tailored to the specific trajectory geometry of the imaging system. This mechanism has been shown to enhance the flexibility and reconstruction efficiency of the method.

The shift-variant FBP algorithm is reformulated into a neural network-compatible representation. This results in the following equation~\cite{ye2024deep,ye2024draco}:
\begin{align*}
x = A_{3d}^T w_{d} A_{2d}^T D w_{red} w_{sino} D A_{2d} w_{cos} p \tag{1}
\label{eq:myequation1}
\end{align*}

In this pipeline, the reconstruction process operates on cone-beam projection data $p$ through a sequence of mathematical operations. These operations include weight applications ($w_{cos}$, $w_{sino}$, $w_{red}$, $w_{d}$), differentiation ($D$), the 2D Radon transform ($A_{2d}$), 2D backprojection ($A_{2d}^T$), and finally, the reconstruction of the volume via 3D backprojection ($A_{3d}^T$). Among these, $w_{red}$ represents trajectory-dependent weights and is the only layer in the reconstruction pipeline that contains learnable parameters. 

While this reconstruction network significantly reduces the number of parameters compared to directly mapping the sinogram to the reconstruction volume with deep neural networks, further compressing the learnable parameters within the network remains a valuable avenue to enhance efficiency.

\subsection{PCA-Based Shift-Variant FBP Model}

\begin{figure}[!t]·
\centering
\includegraphics[width=3in]{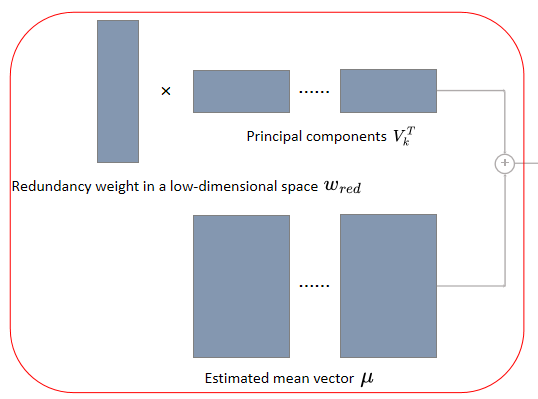}
\caption{PCA Linear Compression Model: Low-dimensional redundancy weights are initially mapped back to the high-dimensional space, and subsequently employed to replace the redundancy weights within the differentiable Shift-Variant FBP neural network.}
\label{PCA}
\end{figure}

Ye et al.~\cite{ye2025compressibilityanalysisdifferentiableshiftvariant} conducted a principal component analysis (PCA) on the redundancy weights derived from training with sinusoidal trajectory data. Their analysis revealed that a limited number of components are sufficient to represent the trajectory-related weight parameters, with negligible impact on reconstruction accuracy. As illustrated in Figure~\ref{PCA}, the redundancy weight layer parameters are decomposed into a trainable eigenvector matrix, compressed weights, and a mean vector.

This decomposition enables the model to learn a compact representation of the weights in a low-dimensional space, while a subsequent transformation maps these representations back to the high-dimensional space required for the reconstruction pipeline. Specifically, the reconstruction of the weights can be expressed as follows:
\begin{align*}
w_{red} = w_{red}' \times V_k^T + \mu,\tag{2}
\label{eq:myequation2}
\end{align*}
where $w_{red}$ represents the reconstructed high-dimensional weight matrix, $V_k$ denotes the eigenvector matrix, $w_{red}'$ represents the compressed weights in the low-dimensional space, and $\mu$ is the mean vector.

Substitution of~\ref{eq:myequation2} into~\ref{eq:myequation1} results in the reconstruction formula for the PCA-based Differentiable Shift-Variant FBP model:
\begin{align*}
x=A_{3d}^Tw_{d}A_{2d}^TD(w_{red}' \times V_k^T + \mu)w_{sino}DA_{2d}w_{cos}p.\tag{3}
\label{eq:myequation10}
\end{align*}

This approach has been demonstrated to markedly reduce the number of trainable parameters while preserving reconstruction accuracy. By optimizing the weight representations in the low-dimensional space, the method enhances training efficiency and reduces model complexity. However, the utilization of linear compression imposes limitations on its functionality, underscoring the imperative for the exploration of more sophisticated compression techniques to attain enhanced performance.

\subsection{2D Gaussian-Based Differentiable Shift-Variant FBP model}
In our previous research, we observed that the model parameters exhibit a simple distribution in high-dimensional space. This characteristic indicates that the spatial arrangement of parameter points can be described by a series of locally concentrated and regular patterns. To achieve a compact representation of this distribution, we employ a 2D GMM for approximation.

\begin{figure}[!t]
\centering
\includegraphics[width=2.1in]{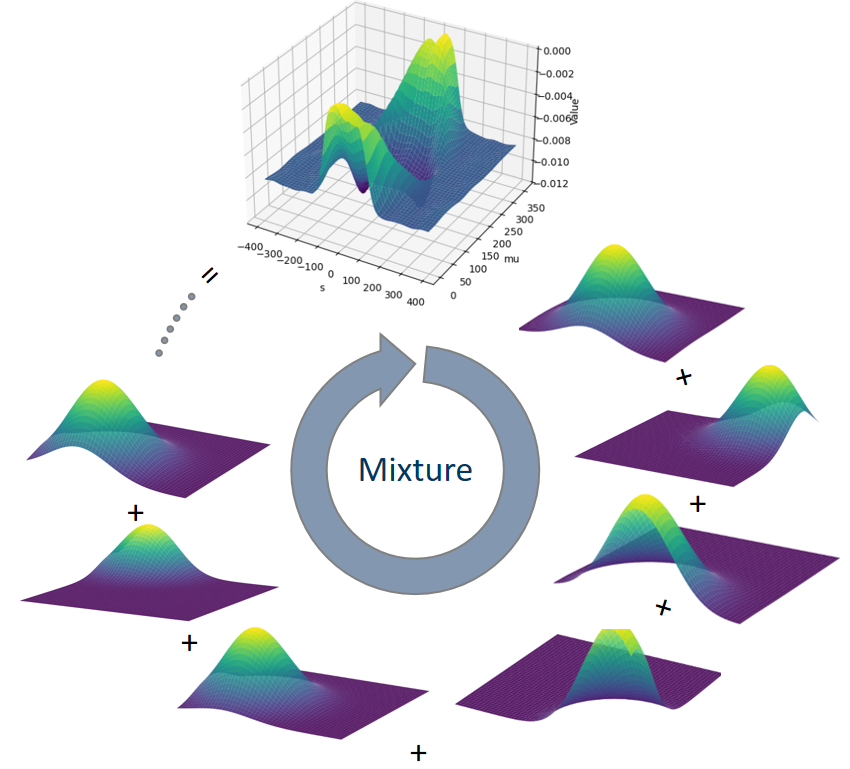}
\caption{2D Gaussian-based Compression Model: Redundancy weights can be interpreted as a linear combination of multiple Gaussian distributions.}
\label{GMM}
\end{figure}

The fundamental principle underlying the GMM involves the decomposition of a complex distribution $p(x, y)$ into a linear combination of multiple Gaussian distributions, as illustrated in Figure~\ref{GMM}, thereby reducing the number of model parameters required to describe the original distribution. To illustrate this concept, consider a set of model parameters defined as $\theta = \{\theta_1,\theta_2,......,\theta_N\}$, with each parameter $\theta_i$ occupying a $M$-dimensional space. Under the framework of GMM, the parameter distribution is modeled as follows:

\begin{align*}
w_{red}=p(x, y) = \sum_{k=1}^K \pi_k \mathcal{N}((x, y) \mid \mu_k, \Sigma_k),\tag{4}
\label{eq:myequation9}
\end{align*}

where $\pi_k$ denotes the weight of the $k-th$ component, and $\mathcal{N}((x, y) \mid \mu_k, \Sigma_k)$ is defined as the multivariate Gaussian distribution with mean $\mu_k$ and covariance matrix $\Sigma_k$.

The employment of GMM is conducive to the effective description of the parameter distribution with a reduced number of variables. This reduction occurs as the original $N\times M$ parameters are transformed into $K\times M $ mean vectors, $K$ mixture weights, and $2K$ diagonal elements of the covariance matrices. This significant reduction not only decreases the storage requirements for the parameters but also preserves their global characteristics.

Substituting~\ref{eq:myequation9} into~\ref{eq:myequation1} yields the formulation of the Gaussian-Based Shift-Variant FBP Neural Network:

\begin{align*}
x=A_{3d}^Tw_{d}A_{2d}^TD(\sum_{k=1}^K \pi_k \mathcal{N}((x, y) \mid \mu_k, \Sigma_k))w_{sino}DA_{2d}w_{cos}p.\tag{5}
\label{eq:myequation11}
\end{align*}

\begin{figure*}[!t]
\centering
\includegraphics[width=7.5in]{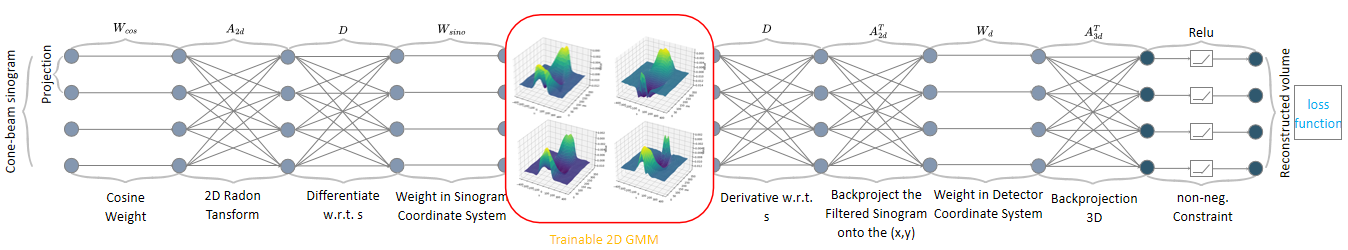}
\caption{GB-SVFBP model: GMM is integrated into the differentiable Shift-variant FBP model to reduce the number of parameters.}
\label{Architecture}
\end{figure*}

Accordingly, the GB-SVFBP model was formulated and subsequently constructed, as illustrated in Figure~\ref{Architecture}.

\subsection{Experiments Setup}
\label{geo}
\subsubsection{Geometry Configuration}
In the present study, the geometric parameters of a clinical C-arm system (specifically, the Artis zeego, manufactured by Siemens AG, Forchheim, Germany) were utilized in the experiment. The detailed geometric parameters are listed in Table~\ref{tab: Geometry Configuration}. The acquisition trajectory employed was a sinusoidal trajectory with a frequency of 5 and a tilt angle range of ±10 degrees. 
\begin{table}[htbp]
\centering
\caption{Geometry configuration}
\resizebox{\columnwidth}{!}{
\begin{tabular}{{@{}cccc@{}}}
\toprule
\textbf{Volume shape} & \textbf{Volume spacing} & \textbf{Number of projections} \\ 
\midrule
128$\times$512$\times$512 & 0.25mm$\times$0.25mm$\times$0.25mm & 400 \\

\toprule
\textbf{Source isocenter distance} & \textbf{Source detector distance}& \textbf{Detector spacing}\\
\midrule
 750mm & 1200mm& 0.616mm$\times$0.616mm \\
 
\toprule
\textbf{Detector shape} &\textbf{Source geometry}   \\ 
\midrule
620$\times$480&Cone beam  \\
\bottomrule
\end{tabular}%
}
\label{tab: Geometry Configuration}
\end{table}

\subsubsection{Data Preparation}



The GB-SVFBP model was trained using the simulation dataset generation method proposed by Ye et al.~\cite{ye2024draco} This method ensures that the neural network effectively learns trajectory-specific weights while avoiding interference from data complexity. Specifically, 30 sets of simulated data were generated, 24 of which were utilized for training and 6 for validation. Subsequently, we employed the forward project module in the Pyro-NN framework~\cite{syben2019pyro} to generate the corresponding sinograms, which served as the input data for the neural network. These sinograms were derived from the trajectory geometry defined in Table~\ref{tab: Geometry Configuration}.

In addition, we utilized the Pancreatic-CT-CBCT-SEG dataset \cite{hong2021breath} to further evaluate the model's performance, which was provided by the Memorial Sloan Kettering Cancer Center and included CT imaging data from 40 patients with pancreatic cancer and generated 20 sets of test data following the same procedure. The sinograms of the test data were also created using the forward project module in Pyro-NN~\cite{syben2019pyro}, while their corresponding ground truth images were directly taken from the original images in the Pancreatic-CT-CBCT-SEG dataset.

By following these steps, we constructed a complete dataset comprising training, validation, and testing data, providing robust support for our experiments.

\subsubsection{Implementation Details}

The implementation of the neural network architecture was conducted using PyTorch 2.1.1, with the 2D Radon transform, 2D inverse Radon transform, and 3D cone-beam backprojection operations facilitated by Pyro-NN~\cite{syben2019pyro}. The loss function incorporated a combination of Mean Squared Error (MSE) and Structural Similarity Index (SSIM) losses, calculated after normalizing the reconstructed and reference results. The model parameters were optimized using the AdamW optimizer with a One-Cycle learning rate policy, varying the learning rate between $0.1$ and $1$ over 1000 epochs. The training was executed on an Nvidia A40 GPU, and the testing was conducted on an Nvidia 4080 GPU.

\section{Results}

\begin{figure}[!t]
\centering
\includegraphics[width=3.5in]{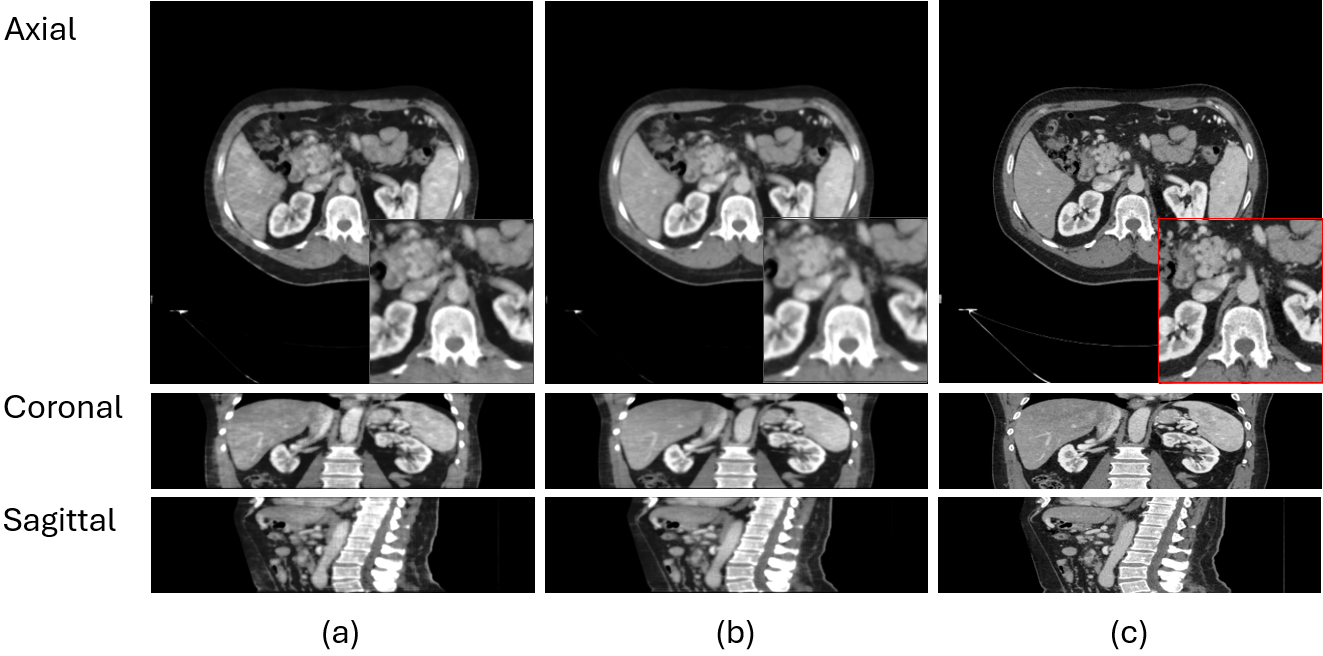}
\caption{Reconstructed results of the network (Sinusoidal orbit). Visualization parameters: window width = 400, window level = 60. (a) Compressed. (b) Uncompressed. (c) Ground truth.}
\label{Results}
\end{figure}

The model was trained on the dataset and achieved convergence after 32 epochs, which required approximately 3 hours. Subsequently, the trained model was utilized for image reconstruction to evaluate its performance. As illustrated in Figure~\ref{Results}, the experimental results indicate that the compressed model exhibits a slight reduction in reconstruction quality compared to the uncompressed model, primarily evident in the loss of image details and the presence of artifacts. Nevertheless, despite the quality degradation, the compressed model still demonstrates high practical usability.


To quantitatively assess the performance of the model, multiple image quality assessment metrics were employed, including MSE, Peak Signal-to-Noise Ratio (PSNR), and SSIM. The performance comparison across different model configurations is summarized in Table~\ref{tab:comparison}. As shown in the table, the uncompressed model achieves the best results across all quality metrics. However, this configuration entails a substantial parameter count of 113,184,000. In contrast, the model compressed using the 2D Gaussian approach significantly reduces the parameter count to 100,000, though this is accompanied by a reduction in image quality.


\begin{table}[htbp]
\centering
\caption{Comparison of Image Quality Metrics}
\resizebox{\columnwidth}{!}{
\begin{tabular}{@{}lcccccc@{}}
\toprule
& \textbf{MSE}$\downarrow$ & \textbf{PSNR (dB)}$\uparrow$ & \textbf{SSIM}$\uparrow$ & \textbf{\#parameters}$\downarrow$ \\ 
\midrule
No compress & 0.0922 ± 0.0119&  38.78± 1.15& 0.9659±0.0057 & 113,184,000\\
PCA-based & 0.0089± 0.0108&  39.05± 1.13& 0.9544±0.0079 & 3,116,560\\
2D Gaussian & 0.1003± 0.0121&  38.04± 1.16& 0.9367±0.0106 & 100,000\\ 
\bottomrule
\end{tabular}%
}
\label{tab:comparison}
\end{table}

\section{Discussion}
The experimental findings demonstrate that the Gaussian-based compression method substantially reduces the number of parameters (from 113,184,000 to 100,000) while causing only a slight degradation in image quality, as indicated by a slight increase in MSE, a decrease in PSNR by 0.74 dB, and a reduction in SSIM by 0.0292. Notably, the reconstructed results of the compressed model remain visually acceptable and demonstrate high practical value in scenarios where resources are limited. This Gaussian-based nonlinear compression method is particularly well-suited for differentiable shift-variant FBP models and does not necessitate additional constraints on the smoothness of the weights.

\section{Conclusion}
This study proposes a Gaussian-Based Shift-Variant FBP neural network that achieves a 99\% reduction in trainable parameters while maintaining competitive reconstruction quality. Additionally, the method significantly accelerates training, reducing training time to one-fourth of the original, which is reduced to a duration of just three hours. The proposed approach demonstrates strong potential for resource-constrained applications. Future work will focus on enhancing reconstruction quality and exploring advanced compression techniques to further optimize the balance between efficiency and performance.

\printbibliography
\end{document}